\documentclass[conference]{IEEEtran}
\IEEEoverridecommandlockouts
\usepackage{cite}
\usepackage{amsmath,amssymb,amsfonts}
\usepackage{algorithmic}
\usepackage{graphicx}
\usepackage{textcomp}
\usepackage{xcolor}
\usepackage{bm}
\usepackage{booktabs}
\usepackage{subfig}
\usepackage{float}
\usepackage[normalem]{ulem}
\usepackage{subcaption}
\usepackage{graphicx}

\def\BibTeX{{\rm B\kern-.05em{\sc i\kern-.025em b}\kern-.08em
    T\kern-.1667em\lower.7ex\hbox{E}\kern-.125emX}}
\begin{document}

\title{Fault-Tolerant Spacecraft Attitude Determination using State Estimation Techniques\\
}

\author{\IEEEauthorblockN{1\textsuperscript{st} Bala Chidambaram}
\IEEEauthorblockA{\textit{Aero/Astro Dept.} \\
\textit{Stanford University}\\
Palo Alto, CA \\
balaguruna.chidambaram@boeing.com}
\and
\IEEEauthorblockN{2\textsuperscript{nd} Anna Hylbert}
\IEEEauthorblockA{\textit{Aero/Astro Dept.} \\
\textit{Stanford University}\\
Palo Alto, CA \\
anna05@stanford.edu}
\and
\IEEEauthorblockN{3\textsuperscript{rd} Miggy Silva}
\IEEEauthorblockA{\textit{Aero/Astro Dept.} \\
\textit{Stanford University}\\
Palo Alto, CA \\
silvam31@stanford.edu}
}

\maketitle

\begin{abstract}

The extended and unscented Kalman filter, and the particle filter provide a robust framework for fault-tolerant attitude estimation on spacecraft. This paper explores how each filter performs for a large satellite in  a low earth orbit. Additionally, various techniques, built on these filters,  for fault detection, isolation and recovery from erroneous sensor measurements, are analyzed. Key results from this analysis include filter performance for various fault modes.

\end{abstract}

\section{Introduction}


Communication satellites, satellites conducting scientific research, and reentry vehicles are all examples of spacecraft that need to predict and control their attitude. Without accurate attitude estimates, vehicles could lose control authority, run out of power due to mis-oriented solar panels, or fail to complete a mission due to improper payload deployment.

Satellite use a suite of sensors, e.g., gyroscopes, magnetometers, and star trackers, in conjunction with filtering techniques like the Extended Kalman Filter (EKF) , the Unscented Kalman filter (UKF) and the Particle Filter (PF), for attitude estimation.  However, these sensor measurements are unreliable, and the harsh environment of space often induces faults. E.g., gyroscope measurements could have a constant bias, and could be contaminated by noise spikes or intermittent dropouts. 

EKFs, UKFs, and PFs, provide a robust framework for Fault Detection, Isolation and Recovery (FDIR). For the purpose of testing and building on these filters, a model of the spacecraft dynamics was created. The model was built using momentum conservation assumptions from rigid body motion, and the assumption that the only significant perturbation on the system was gravity gradient torque. Quaternion representations were used for the attitude and the  performance of the filters was analyzed.

A variety of techniques for FDIR were then incorporated into the filters. KF innovation residual statistics such as innovation filtering and innovation sequence monitoring detected both noise spikes and intermittent drop-outs. Variations of these techniques were also used for fault isolation. The KF formulations provide implicit fault recovery and state augmentation works well for bias estimation. The validity of these techniques were then analyzed by comparing their perfomance to that of standard EKFs, UKFs, and PFs.

Section \ref{Related_work} surveys related past work.  Section \ref{tech_approach} describes the satellite, sensor and fault models, and develops the FDIR techniques. In Section \ref{results_discussion}, several experiments are run exercising the techniques and discussing the results. Finally,  key insights are summarized in Section \ref{conclusion}.

\section{Related work} \label{Related_work}



\subsection{Common Sensors and Attitude Representations}

Gyroscopes measure the angular velocity along a given axis. However, the measurements include a constant bias. \cite{survey_nonlinear_filters} Common attitude sensors include sun sensors, star trackers, and magnetometers. In high precision models, star tracker measurements can be modeled as vectors pointing to different stars that need to be converted into the spacecraft's body frame to determine the attitude. \cite{foundational_ekf_nasa} In practice, quaternions are useful for expressing the attitude of a spacecraft due to them avoiding singularities that exist in Euler Angles that can result in the loss of control authority of a spacecraft. However, Euler angles are still used as references due to their direct physical interpretation for analysis. \cite{pf_without_gyro}.

\subsection{Foundational EKF, UKF, and PF}

The EKF is a widely used recursive estimator for nonlinear systems due to it using a fairly straightforward approach of linearizing about the current best estimate of the problem. \cite{survey_nonlinear_filters} While the EKF performs effectively under mild nonlinearities and small initial errors, the process of linearizing the system becomes ineffective when faced with highly nonlinear dynamics.\cite{quest_model}

Due to the EKF truncating all but the first-order terms when linearizing the system, the performance can be limited. Through using the unscented transformation (UT), the UKF can capture these higher order terms in the nonlinear system and therefore lead to better performance. \cite{ukf_spacecraft_attitude_vd_cs} 



As opposed to the EKF and UKF, PFs don't rely on the underlying distribution being Gaussian. When designing a PF, it is deciding how many times particles should be resampled so that they're weighted accordingly. Oversampling can lead to particle impoverishment and undersampling can lead to particle degeneracy. \cite{survey_nonlinear_filters} \cite{pf_without_gyro}







\subsection{Faulty Sensor Measurements} \label{sec:faulty_sensor_techniques}

In a federated UKF architecture, each local UKF processes measurements from an individual sensor, and their filtered estimates are then fused by a central filter. This enhances fault tolerance by isolating sensor-specific errors and allows the system to gracefully degrade or ignore fault data without corrupting the entire state estimate.\cite{actuator_sensor_faults}


The PF can be used to perform hypothesis testing on the health of the sensor, across various operational modes. \cite{wright_pf_detect_faulty} Similarly, EKF and UKF algorithms where the Kalman gains are altered if there are step changes in measurements can be used to safely weight them. \cite{Yung_detect_twostage} In another approach, known sensor parameters such as bias can be corrected through augmenting them into the EKF state. \cite{robust_kalman_small_satellite}. All of these methods are summarized nicely in the framework of FDIR that relies on concepts such as innovation filtering and sequencing for fault detection, innovation filtering and sequencing for fault isolation, and redundant sensor fusion and bias estimation for fault recovery. \cite{groves_principles}

\section{Technical Approach} \label{tech_approach}

To simulate the noiseless measurements of the spacecraft, the attitude and angular velocity were propagated by assuming that gravity was the only force acting on the system. Attitude measurements from a star tracker and magnetometer were then simulated through taking the attitude ground truths and adding gaussian noise with covariance values expected from each sensor. Gyroscope measurements were simulated by adding gaussian noise to the angular velocities as well as a constant drift value. The measurements were then fed into an EKF, UKF, and PF to estimate the state of the system.

The next stage was subjecting the sensors to various faults including noise spikes, intermittent dropouts and biases. Filters build on state estimation provide a robust set of techniques for sensor FDIR. Statistics from these filters were used to detect and isolate both transient and persistent faults. Finally sensor fusion and state augmentation techniques were used for recovery from faults.

\subsection*{Notation}

\begin{itemize}
  \item $\omega_{x,k}, \omega_{y,k}, \omega_{z,k}$: angular velocity components
  \item $\tau_{x,k}, \tau_{y,k}, \tau_{z,k}$: torques applied about each axis
  \item $b_{x,k}, b_{y,k}, b_{z,k}$: gyroscope biases
  \item $\theta_{x,k}, \theta_{y,k}, \theta_{z,k}$: Euler angles (roll, pitch, and yaw)
  \item $q_{0,k}, q_{1,k}, q_{2,k}, q_{3,k}$: attitude quaternion
  \item $I_{xx}, I_{yy}, I_{zz}$: principal moments of inertia
\end{itemize}

\subsection{Dynamics Model}
\label{sec:dynamics_model}

\subsubsection{State Vector}
\label{state_vector}

Euler angles and quaternions complement each other well in state estimation. Euler angles provide physical intuition for the system, but quaternions tend to work better for filtering for reasons discussed in Section \ref{Related_work}. Therefore, two state vectors were propagated.
\[
\bm{x} =
\begin{bmatrix}
\theta_{x} & \theta_{y} & \theta_{z} &
\omega_{x} & \omega_{y} & \omega_{z} &
\end{bmatrix}^\top
\in \mathbb{R}^6
\]
\[
\bm{x} =
\begin{bmatrix}
q_{0} & q_{1} & q_{2} & q_{3} &
\omega_{x} & \omega_{y} & \omega_{z} &
\end{bmatrix}^\top
\in \mathbb{R}^7
\]
The continuous derivative functions defined in Sections \ref{ang_vel} through \ref{quat} were fed into a RK45 ODE solver to obtain ground truth values at each discrete timestep $k$.

\subsubsection{Angular Velocity Rates} \label{ang_vel}

The angular velocity was updated using Euler's Equation for rigid body motion. \noindent
\begin{align} 
\dot\omega_{x} &= \frac{1}{I_{xx}} \left[ \tau_{x} - (I_{zz} - I_{yy}) \omega_{y} \omega_{z} \right] \label{eq:omega_x} \\
\dot\omega_{y} &= \frac{1}{I_{yy}} \left[ \tau_{y} - (I_{xx} - I_{zz}) \omega_{z} \omega_{x} \right] \label{eq:omega_y} \\
\dot\omega_{z} &= \frac{1}{I_{zz}} \left[ \tau_{z} - (I_{yy} - I_{xx}) \omega_{x} \omega_{y} \right] \label{eq:omega_z}
\end{align}

\subsubsection{Euler Angles} \label{euler}

A 313 Euler sequence (rotating about the z, then x, then z axes) was used to propagate the Euler angles from the angular velocities. Equation \ref{eq:euler_update} accounts for the modified coordinate system between each of the subsequent rotations where $s()$ and $c()$ represent sin and cosine respectively.

\begin{align}  \label{eq:euler_update}
\begin{bmatrix}
    \dot\phi \\
    \dot\theta \\
    \dot\psi \\
\end{bmatrix}
= 
\frac{1}{s(\theta)}
\begin{bmatrix}
    -s(\phi)c(\theta) & -c(\phi)c(\theta) & s(\theta) \\
    c(\phi)s(\theta) & -s(\phi)s(\theta) & 0 \\
    s(\phi) & c(\phi) & 0
\end{bmatrix}
\begin{bmatrix}
    \omega_x \\
    \omega_y \\
    \omega_z \\
\end{bmatrix}
\end{align}

\subsubsection{Quaternions} \label{quat}

Equation \ref{eq:quat_update} updates quaternions with the angular velocity while preserving their magnitude constraint.

\begin{align}  \label{eq:quat_update}
\begin{bmatrix}
    \dot q_0 \\
    \dot q_1 \\
    \dot q_2 \\
    \dot q_3
\end{bmatrix}
= 
\frac{1}{2}
\begin{bmatrix}
    -q_1 & q_0 & q_3 & -q_2 \\
    -q_2 & -q_3 & q_0 & q_1 \\
    -q_3 & q_2 & -q_1 & q_0
\end{bmatrix}'
\begin{bmatrix}
    \omega_x \\
    \omega_y \\
    \omega_z \\
\end{bmatrix}
\end{align}

\subsubsection{Gravity Gradient Torque}

To model gravity gradient torque, the spacecraft's distance from Earth and orientation in the RTN (radial-transverse-normal) frame needed to be obtained. Equations \ref{eq:grav_grad} and \ref{eq:grav_grad_cs} show the relationship between the gravity gradient torque and the location and orientation of the spacecraft where $\boldsymbol{R}_1$ and $\boldsymbol{R}_2$ are the rotation matrices from the ECI to body frames and ECI to RTN frames respectively.

\begin{equation} \label{eq:grav_grad}
    \boldsymbol{M} = \frac{3\mu_{grav}}{R^3}
    \begin{bmatrix}
        (I_{zz} - I_{yy})c_yc_z \\
        (I_{xx} - I{zz})c_zc_x \\
        (I_{yy} - I_{xx})c_xc_y
    \end{bmatrix}
\end{equation}

\begin{equation} \label{eq:grav_grad_cs}
    \boldsymbol{c} = \boldsymbol{R}_{1} \boldsymbol{R}_{2}^T
    \begin{bmatrix}
        1 & 0 & 0
    \end{bmatrix}^T
\end{equation}



\subsection{Measurement Model}

\subsubsection{Measurement Matrix}

Due to having dual attitude measurements from the star tracker and magnetometer, the measurement vector had two forms for Euler angles and quaternions.

\begin{align}
\begin{split}
\bm{y} &= 
\bigl[\,y^{\text{st}}_{x}\;y^{\text{st}}_{y}\;y^{\text{st}}_{z}\;
       y^{\text{mm}}_{x}\;y^{\text{mm}}_{y}\bigr.\\
     &\quad\bigl.y^{\text{mm}}_{z}\;y^{\text{gyro}}_{x}\;
               y^{\text{gyro}}_{y}\;y^{\text{gyro}}_{z}\bigr]^\top
\;\in\;\mathbb{R}^{9}
\end{split} \\
\begin{split}
\bm{y} &= 
\bigl[\,y^{\text{st}}_{0}\;y^{\text{st}}_{1}\;y^{\text{st}}_{2}\;
       y^{\text{st}}_{3}\;y^{\text{mm}}_{0}\;y^{\text{mm}}_{1}\bigr.\\
     &\quad\bigl.y^{\text{mm}}_{2}\;y^{\text{mm}}_{3}\;y^{\text{gyro}}_{x}\;
               y^{\text{gyro}}_{y}\;y^{\text{gyro}}_{z}\bigr]^\top
\;\in\;\mathbb{R}^{11}
\end{split}
\end{align}

\subsubsection{Gyroscope} The Gyroscope directly measures angular velocity along each principle axis.

\begin{align}
y^{\text{gyro}}_{x,k} &= \omega_{x,k} + b_{x,k} + v_{x,k} \label{eq:meas_x}\\
y^{\text{gyro}}_{y,k} &= \omega_{y,k} + b_{y,k} + v_{y,k} \label{eq:meas_y}\\
y^{\text{gyro}}_{z,k} &= \omega_{z,k} + b_{z,k} + v_{z,k} \label{eq:meas_z}
\end{align}

In Equations \ref{eq:meas_x}, \ref{eq:meas_y}, and \ref{eq:meas_z}, $v_{x,k}, v_{y,k}, v_{z,k} \sim \mathcal{N}(0, \sigma^2)$. 

\subsubsection{Star Tracker and Magnetometer} Star trackers and magnetometers measure the attitude of the spacecraft. For both Euler angles and quaternions, each sensor provides a measurement of the ground truth value plus noise sampled from a gaussian distribution: $n_k^{st} \sim \mathcal{N}(0,\sigma_{\text{st}}^2)$ and $n_k^{mm} \sim \mathcal{N}(0,\sigma_{\text{mm}}^2)$.



\subsection{EKF, UKF, and PF Formulations}
\label{KF_Formulations}




\subsubsection{Notation}
$x_t$: state, $\mu_t = \mathbb{E}[x_t]$, $\Sigma_t = \text{Cov}[x_t]$, 
$u_t$: control input, $y_t$: measurement, 
$f(\cdot)$: dynamics, $h(\cdot)$: measurement, 
$Q$, $R$: noise covariances, 
$A_t = \partial f / \partial x$, $H_t = \partial h / \partial x$, 
$\chi^{(i)}$: sigma points, $x_t^{[i]}$: particles, $w^{(i)}$: weights

\subsubsection{Extended Kalman Filter (EKF)} \label{subsub:EKF}

\small{
\begin{equation}
\begin{aligned}
\textit{Prediction:} \quad
& \mu^- = f(\mu, u), \\
& \Sigma^- = A \Sigma A^\top + Q \\
\textit{Update:} \quad
& K = \Sigma^- H^\top (H \Sigma^- H^\top + R)^{-1}, \\
& \mu = \mu^- + K(y - h(\mu^-)), \\
& \Sigma = (I - K H) \Sigma^-
\end{aligned}
\end{equation}
}

\subsubsection{Unscented Kalman Filter (UKF)}
The UKF formulation is shown below:

\small{
\begin{equation}
\begin{aligned}
\textit{Prediction:} \\
\chi^{(i)} &= \text{SigmaPts}(\mu, \Sigma), \quad
\chi^{-(i)} = f(\chi^{(i)}, u) \\
\mu^- &= \sum_i w_m^{(i)} \chi^{-(i)}, \\ 
\Sigma^- &= \sum_i w_c^{(i)} (\chi^{-(i)} - \mu^-)(\cdot)^\top + Q \\
\textit{Update:} \\
\zeta^{(i)} &= h(\chi^{-(i)}), \quad
\hat{y} = \sum_i w_m^{(i)} \zeta^{(i)} \\
K &= \sum_i w_c^{(i)} (\chi^{-(i)} - \mu^-)(\zeta^{(i)} - \hat{y})^\top \\
& \quad \cdot \left[\sum_i w_c^{(i)} (\zeta^{(i)} - \hat{y})(\cdot)^\top + R\right]^{-1} \\
\mu &= \mu^- + K(y - \hat{y}), \quad
\Sigma = \Sigma^- - K(\cdot)K^\top
\end{aligned}
\end{equation}
}

\subsubsection{Particle Filter (PF)}
The PF represents the belief over the state \( x_t \) using a set of \( N\) weighted samples \( \{x_t^{[i]}, w_t^{[i]}\}_{i=1}^N \). It propagates the posterior distribution through a sequence of sampling and resampling steps:

\small{
\begin{equation}
\begin{aligned}
\textit{Prediction:} \quad & x^{[i]} \sim p(x \mid x^{[i]}, u) \\
\textit{Update:} \quad & w^{[i]} \propto p(y \mid x^{[i]}), \quad
w^{[i]} \leftarrow \frac{w^{[i]}}{\sum_j w^{[j]}} \\
\textit{Resample:} \quad & x^{[i]} \sim \text{Resample}(\{x^{[j]}, w^{[j]}\})
\end{aligned}
\end{equation}
}







\subsection{Sensor Fault Detection, Isolation and Recovery }
\label{FDIR}

Typical faults for gyroscopes, star-trackers and magnetometers include noise spikes, intermittent dropouts, biases, saturation, etc.
Table \ref{FDIR_table} summarizes FDIR techniques for these faults. The filters were then used to develop models to implement these techniques.

\begin{table}[t]
\centering
\caption{Sensor FDIR with Kalman Filtering Techniques}
\label{FDIR_table}
\begin{tabular}{|c|c|c|}
\hline
\textbf{Fault Type} & \textbf{Detection/Isolation} & \textbf{Recovery} \\
\hline
Constant Bias & Automatic & Bias Estimation \\
Noise Spike & Innovation  Filtering & Sensor Fusion \\
Step Fault & Innovation Sequence Monitoring & Sensor Fusion \\
Saturation & Inno. Filtering & Sensor Fusion\\
\hline
\end{tabular}
\end{table}





\subsubsection{Fault Detection}
\label{IFISM}
Fault detection methods using Kalman Filter statistics include innovation filtering and innovation sequence monitoring.

\textbf{Innovation Filtering} detects sudden, anomalous deviations in the measurement model at a single time step making it effective for identifying outliers, spikes or saturation events.

\textbf{Innovation Sequence Monitoring} detects gradual or persistent sensor faults by analyzing trends or cumulative behavior in the innovation history over time making it effective for identifying slow drifts, calibration shifts, or systematic bias.

The mathematical formulations of these techniques are now explored.

\paragraph{Innovation Filter Model}

 For the Innovation Filter (EKF formulation), the innovation at time \( t \) is given by:
\begin{equation}
    \nu_t = y_t - C \mu_{t|t-1} - D u_t
\end{equation}

The corresponding innovation covariance is:
\begin{equation}
    S_t = C \Sigma_{t|t-1} C^\top + R
\end{equation}

The normalized innovation squared (NIS) is:
\begin{equation}
\label{eq:NIS_EKF}
    \text{NIS}_t = \nu_t^\top S_t^{-1} \nu_t
\end{equation}

Under nominal conditions, \( \text{NIS}_t \sim \chi^2_k \), where \( k \) is the dimension of the measurement vector (we assume a 0.95 confidence level). A fault is detected when the following condition is met:
\begin{equation}
\label{NIS_Threshold}
    \text{NIS}_t > \gamma \quad \Rightarrow \quad \text{sensor anomaly detected}
\end{equation}

For the Innovation Filter (UKF formulation),given the sigma points, weights, and measurements, the prediction is give  by:
\begin{equation}
\hat{y} = \sum_{i} w_i \, y_i
\end{equation}

The measurement covariance is given as follows:
\begin{equation}
\label{sigma_yy}
\Sigma_{yy} = \sum_{i} w_i (y_i - \hat{y})(y_i - \hat{y})^T + R
\end{equation}

The NIS statistics is calculated and applied as before (Eq.\ref{eq:NIS_EKF}), but using $\Sigma_{yy}$.

PFs don't have an innovation vector or an innovation covariance matrix in closed form, but NIS-like consistency checks can still be used.

The predicted measurement mean is

\[
\hat{y}_t = \sum_{i=1}^N w_t^{[i]} g(x_t^{[i]})
\]

The predicted measurement covariance is
\[
S_t = \sum_{i=1}^N w_t^{[i]} \left( g(x_t^{[i]}) - \hat{y}_t \right) \left( g(x_t^{[i]}) - \hat{y}_t \right)^\top
\]
Again, the NIS statistics is calculated and applied as before

\paragraph{Innovation Sequence Filter Model}
For the Innovation Sequence filter (EKF), the average NIS (moving average) is given by

\begin{equation}
\label{seq_threshold}
    \overline{\text{NIS}}_t = \frac{1}{N} \sum_{i=t-N+1}^{t} \nu_i^\top S_i^{-1} \nu_i
\end{equation}

where \( N \) is the window size, and \( \nu_i \) and \( S_i \) are the innovation and innovation covariance at time \( i \), respectively.

Under nominal conditions, \( \overline{\text{NIS}}_t \sim \chi^2_k \). A fault is detected when the following condition is met:
\begin{equation}
\label{NIS_threshold_seq}
    \overline{\text{NIS}}_t > \gamma \quad \Rightarrow \quad \text{sensor anomaly detected}
\end{equation}

The derivations of the innovation filters for the UKF and the PF are similar to what was shown in Section \ref{IFISM}.

\subsubsection{Fault Isolation}
\label{FIIF}

Fault detection can easily be applied to fault isolation, i.e., identifying the specific sensor that is faulty by using a per sensor version of Eqs. \ref{NIS_Threshold} and \ref{seq_threshold}.

\begin{itemize}
    \item \textbf{Innovation Filtering} detects the sensor experiencing the spike by computing ${\text{NIS}}_i$, extracting the measurements, measurement Jacobian and noise covariance corresponding to each sensor i. The $\gamma$ for the desired confidence level $\alpha$ is then computed but with the reduced degrees of freedom $d$, corresponding to the sensor measurements
    \item \textbf{Innovation Sequence Monitoring}  detects the persistent sensor fault a similar way.
\end{itemize}


The fault isolation also provide a mechanism to slice the measurements, measurement Jacobian and noise covariance to only the valid sensor entries, i.e., the ones where the $\text{NIS}$ is less than the threshold. The provides techniques for recovery which will be the focus of the next section.

\subsubsection{Recovery -  Redundant Sensor Fusion and Bias Estimation}
\label{sec:recovery}

Once a fault is detected and isolated, the state estimate system is needed to continue to provide reliable state estimates. Redundant sensor fusion and bias estimation are two powerful techniques that the filters provide in this context.

\textbf{Redundant Sensor Fusion } excludes the faulty sensor from the measurement update and uses the remaining healthy sensors to continue the estimation process. The filters exhibit an inherent fault tolerance with redundant sensor fusion.

\textbf{Bias estimation } incorporates faults such as constant biases as part of the state vector and then performs the estimation. The filter can thus compensate for systematic errors rather than discard the sensor value completely.

The mathematical formulations of these techniques are now derived.

\paragraph{Redundant Sensor Fusion}     Once the fault is isolated as discussed in Section \ref{FIIF}, a simplified KF formulation is obtained. E.g., if the gyroscope is found to be faulty, valid matrices are defined.

\begin{equation}
\begin{split}
y_t^{(\text{valid})} &= \begin{bmatrix}
y_{\text{mag}} \\
y_{\text{star}}
\end{bmatrix}, \quad
H_t^{(\text{valid})} = \begin{bmatrix}
H_{\text{mag}} \\
H_{\text{star}}
\end{bmatrix}, \quad \\
R_t^{(\text{valid})} &= \mathrm{blkdiag}(R_{\text{mag}}, R_{\text{star}})
\end{split}
\end{equation}

The EKF Kalman filter update equations in Section \ref{subsub:EKF} can then be used with these valid matrices.

\paragraph{Bias Estimation Model}; The state vector described in Section \ref{state_vector} is then augmented with the gyroscope bias as shown below where $b_{x}, b_{y}, b_{z} $ are the gyroscope biases.

{\scriptsize
\begin{equation}   
\bm{x} =
\begin{bmatrix}
q_{0} & q_{1} & q_{2} & q_{3} &
\omega_{x} & \omega_{y} & \omega_{z} &
 b_{x} & b_{y} & b_{z} 
\end{bmatrix}^\top
\in \mathbb{R}^{10}
\end{equation}
}



\section{Results and Discussion}
\label{results_discussion}
\subsection{Baseline Satellite Model Parameters}

The initial values for the Euler angles, quaternions, and angular velocities were as follows: $ \omega_0 = \begin{bmatrix} -7 & 2 & 5 \end{bmatrix}^T $ deg/s, $\alpha_0 = \begin{bmatrix} 0.5 & 1.0 & 1.5 \end{bmatrix}^T $ deg, $q_0 = \begin{bmatrix} 1 & 0 & 0 & 0 \end{bmatrix}^T $.

The physical and orbital parameters for the spacecraft were chosen to loosely align with a large satellite in a nearly circular, polar orbit:     $a = 7080.6$ km, $e = 0.0000979$, $i = 98.2$ deg, $\omega = 120.4799$ deg, $\Omega = 95.2063$ deg, $\nu_0 = 0$ deg,

{\small
\begin{align*}
    I_p &= \begin{bmatrix}
       23745 &93.907& -1267.1\\
    93.907 &17560 &-967.50\\
    -1267.1 &-967.5 &36065
    \end{bmatrix} \\
\end{align*}
}

Finally, the covariance matrices for the sensors were chosen to align with their expected precision and the attitude parameterization size: 
    $\Sigma_{st}^{ea} = 0.001\boldsymbol{I}_3$, $\Sigma_{st}^{quat} = 0.001\boldsymbol{I}_4$, $\Sigma_{mm}^{ea} = \begin{bmatrix} 0.01 & 0.02 & 0.05 \end{bmatrix}$, $\Sigma_{mm}^{quat} = \begin{bmatrix}     0.01 & 0.02 & 0.05 & 0.03 \end{bmatrix}$, $\sigma_{gyro} = 0.005$, $bias_{gyro} = \begin{bmatrix} 0.02&-0.015&0.01 \end{bmatrix}$.

\subsection{Simulation Results}

\begin{figure}[htbp]
    \centering
    \subfloat[Euler Angles vs Time, Ground Truth]{\includegraphics[width=0.45\linewidth]{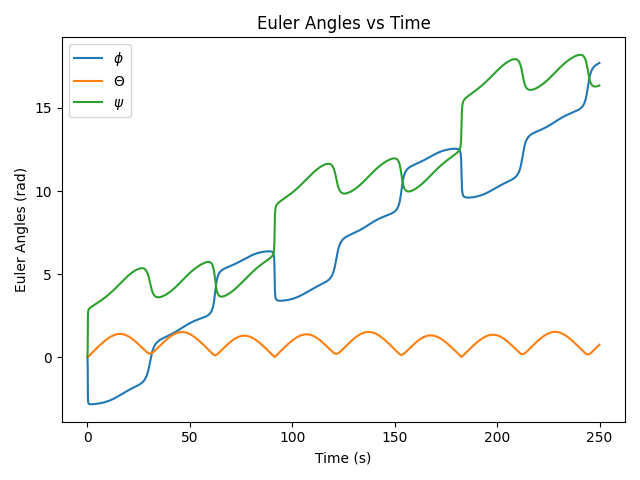}} \hfill
    \subfloat[Quaternions vs Time, Ground Truth]{\includegraphics[width=0.45\linewidth]{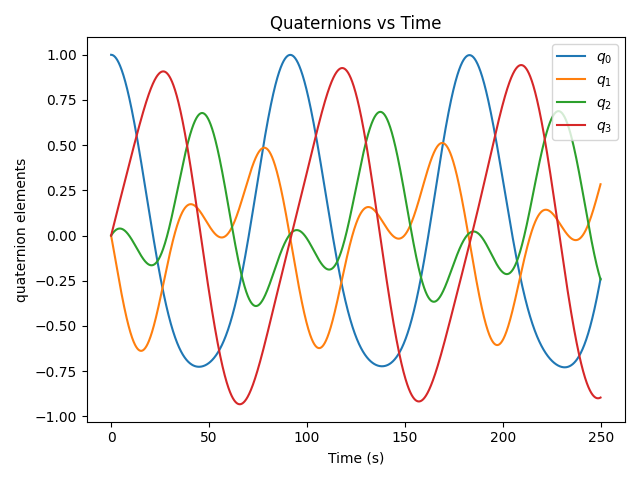}} \\[1mm]
    \subfloat[Angular Velocity vs Time, Ground Truth]{\includegraphics[width=0.45\linewidth]{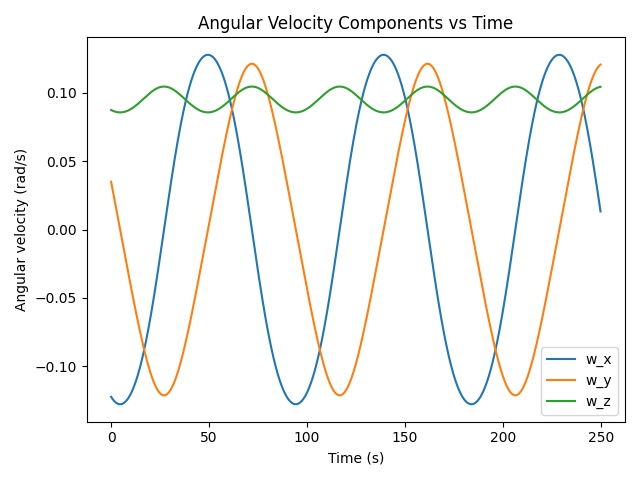}} \hfill
    \subfloat[Orbit]{\includegraphics[width=0.45\linewidth]{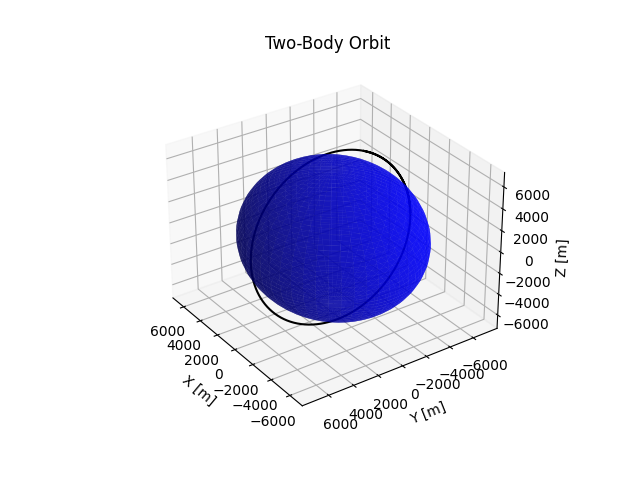}}
    \caption{Simulation Ground Truth}
    \label{fig:sim_results}
\end{figure}

Propagating the angular velocity, Euler angles, quaternions, and orbit resulted in Figure \ref{fig:sim_results}. 

\subsection{EKF, UKF and PF Results}

\subsubsection{Euler Angles vs. Quaternions}

\begin{figure}[htbp]
    \centering
    \subfloat[UKF Euler Angles vs. Time, Ground Truth]{\includegraphics[width=0.45\linewidth]{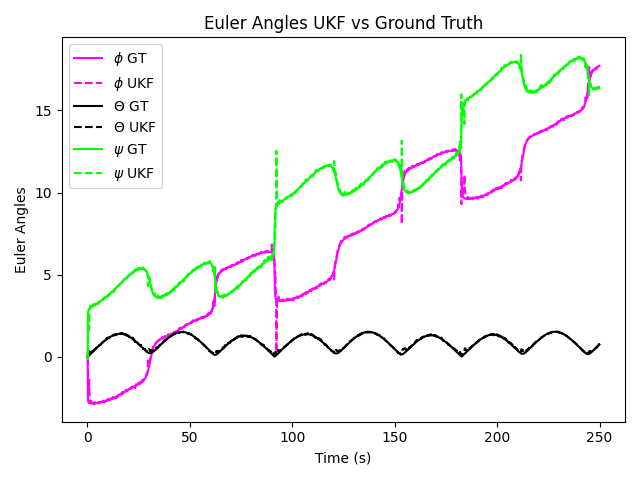}} \hfill
    \subfloat[UKF Euler Angles Error vs Time]{\includegraphics[width=0.45\linewidth]{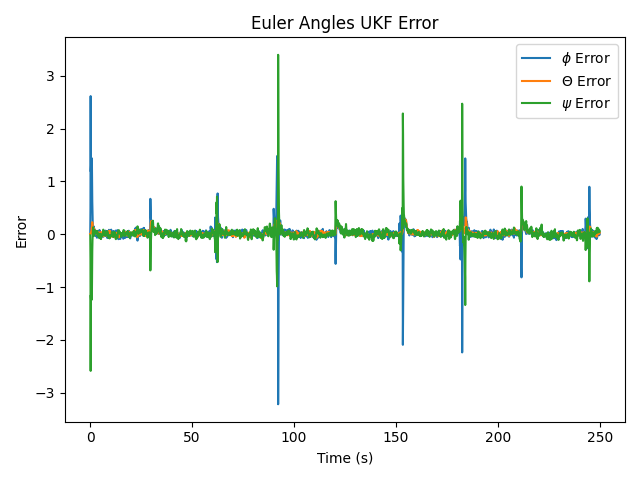}}
    \caption{UKF Euler Angle Estimate}
    \label{fig:euler_ukf}
\end{figure}

The steep inclines that happen approximately every 30 seconds in the Euler angle propagation in Figure \ref{fig:sim_results} led to spikiness in both the UKF and EKF. An example of this spikiness from the UKF is shown in Figure \ref{fig:euler_ukf}. Due to this, quaternions were used for the remaining analysis in this paper.

\subsubsection{Gravity Gradient Analysis}

\begin{figure}
    \centering
    \includegraphics[width=0.70\linewidth]{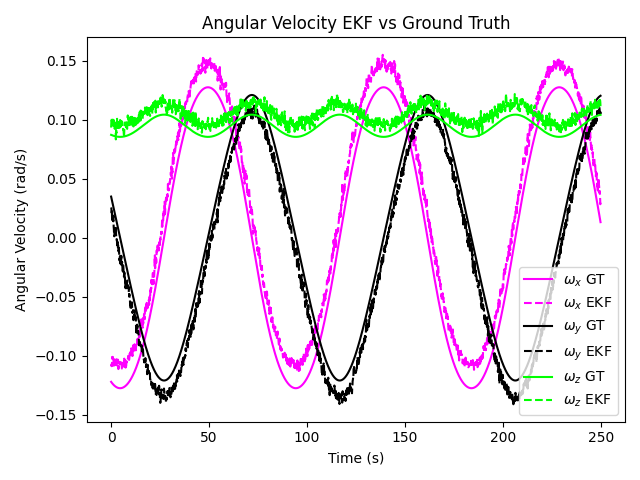}
    \caption{EKF Angular Velocity vs Time without Gravity Gradient Effect}
    \label{fig:ekf_grav_grad}
\end{figure}

\begin{figure}
    \centering
    \includegraphics[width=0.70\linewidth]{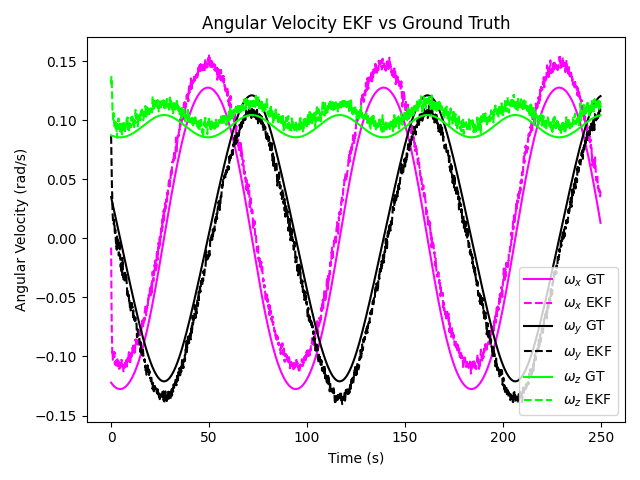}
    \caption{EKF Angular Velocity vs Time with Gravity Gradient Effect}
    \label{fig:ekf_grav_grad_ang_vel}
\end{figure}

Adding gravity gradient torque to the simulation provided a framework for analyzing how well the UKF and EKF handled dynamics that were not included in their dynamics functions. For both filters, the error increased significantly for the first few seconds of the simulation before the filter was able to compensate. Results from the EKF are shown in Figures \ref{fig:ekf_grav_grad} and \ref{fig:ekf_grav_grad_ang_vel}.


\subsubsection{EKF Results}

\begin{figure}
    \centering
    \includegraphics[width=0.70\linewidth]{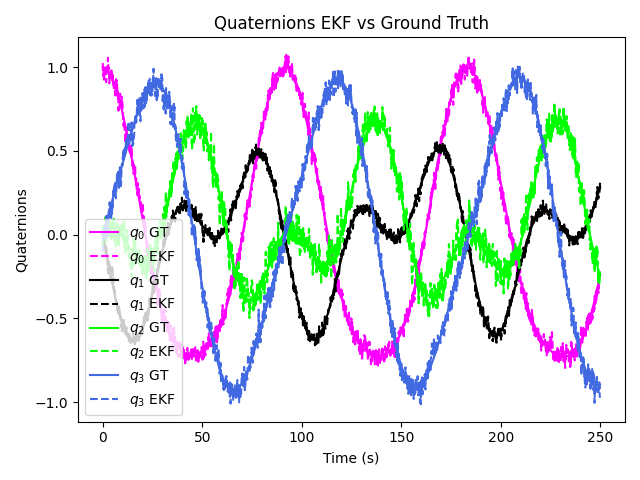}
    \caption{EKF Quaternions vs Time, Estimate compared to Ground Truth}
    \label{fig:ekf_results_quats}
\end{figure}

\begin{figure}
    \centering
    \includegraphics[width=0.70\linewidth]{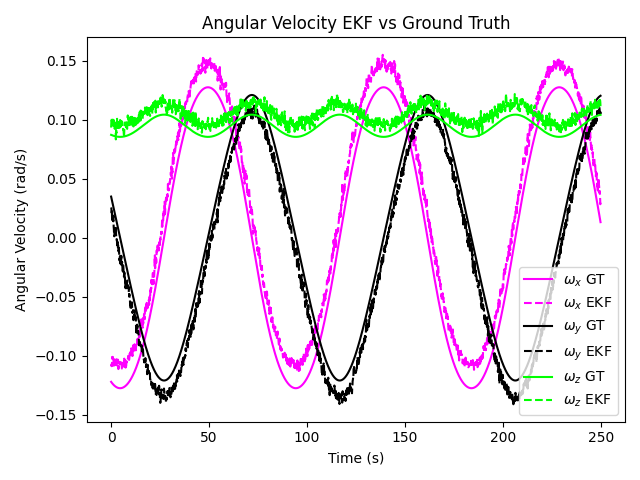}
    \caption{EKF Angular Velocity vs Time, Estimate compared to Ground Truth}
    \label{fig:ekf_results_ang_vel}
\end{figure}

As shown in Figures \ref{fig:ekf_results_quats} and \ref{fig:ekf_results_ang_vel}, the EKF estimated the attitude of the spacecraft extremely well. However, it wasn't able to correct the bias in the gyroscope measurement.


\subsubsection{UKF Results}

\begin{figure}
    \centering
    \includegraphics[width=0.70\linewidth]{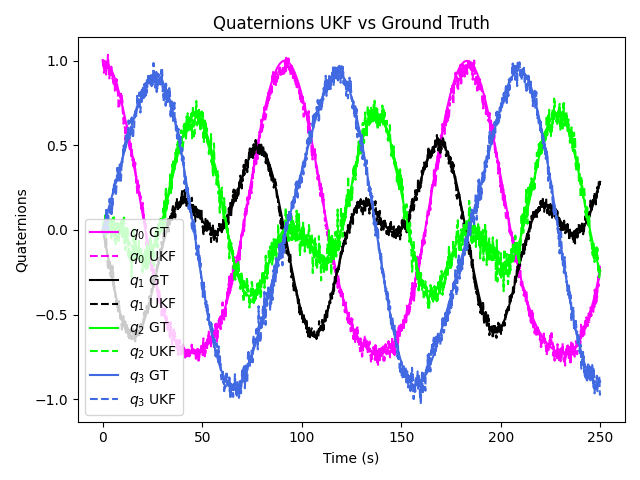}
    \caption{UKF Quaternions vs Time, Estimate compared to Ground Truth}
    \label{fig:ukf_results_quats}
\end{figure}

\begin{figure}
    \centering
    \includegraphics[width=0.70\linewidth]{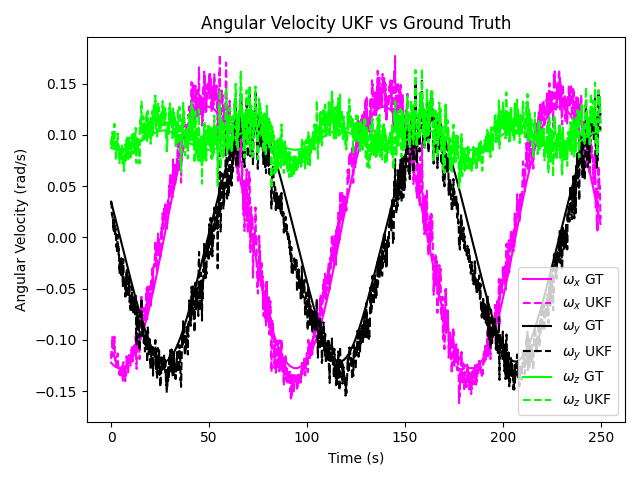}
    \caption{UKF Angular Velocity vs Time, Estimate compared to Ground Truth}
    \label{fig:ukf_results_ang_vel}
\end{figure}

While the noise in the results for the UKF was larger than the EKF, it still estimated the attitude with a low level of error, and it was able to correct the gyroscope bias that the EKF was not. These results are shown in Figures \ref{fig:ukf_results_quats} and \ref{fig:ukf_results_ang_vel}.


\subsubsection{PF Results}
\begin{figure}
    \centering
    \includegraphics[width=0.70\linewidth]{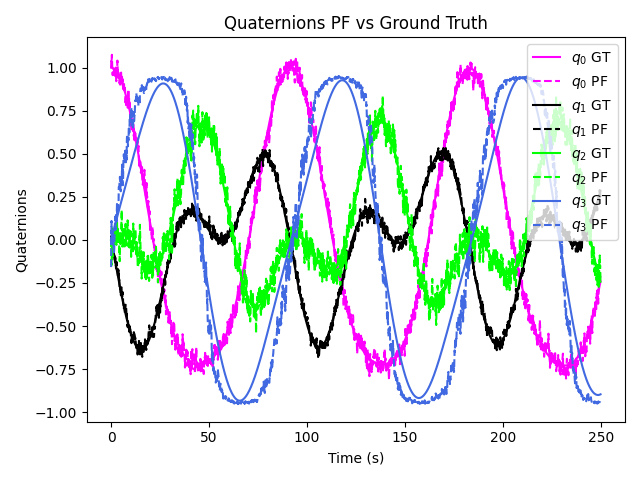}
    \caption{PF Quaternions vs Time, Estimate compared to Ground Truth}
    \label{fig: PF results quaternions}
\end{figure}

\begin{figure}
    \centering
    \includegraphics[width=0.70\linewidth]{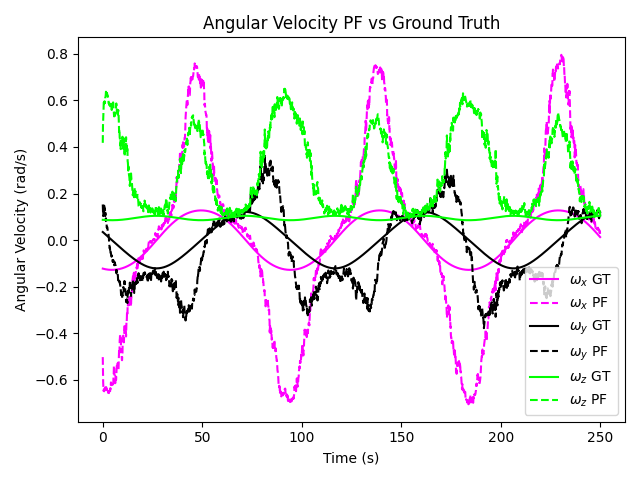}
    \caption{PF Angular Velocity vs Time, Estimate compared to Ground Truth}
    \label{fig: PF results angular velocity}
\end{figure}

As shown in Figures \ref{fig: PF results quaternions} and \ref{fig: PF results angular velocity}, the PF was able to track the attitude of the satellite with low error, but did introduce more error than the EKF and UKF. Additionally, the PF wasn't able to track the angular velocity with a high accuracy.


\vspace{1mm}

\subsection{FDIR Experiments and Results}

\subsubsection{Sensor Fault Detection}
\label{FI_1}

\begin{figure}
    \centering
    \includegraphics[width=0.70\linewidth]{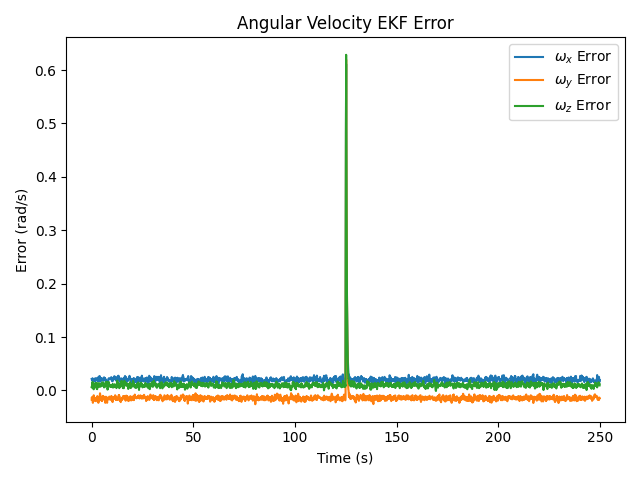}
    \caption{EKF Angular Velocity Error vs Time, without Innovation Filtering}
    \label{fig: EKF results without innovation filtering}
\end{figure}

\begin{figure}
    \centering
    \includegraphics[width=0.70\linewidth]{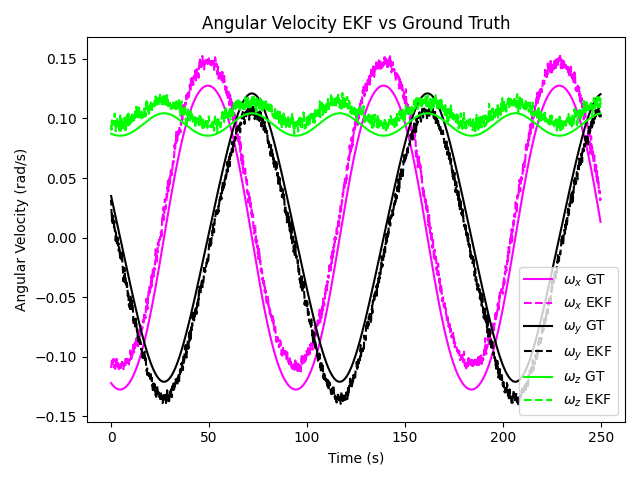}
    \caption{EKF Angular Velocity vs Time, with Innovation Filtering}
    \label{fig: EKF results with innovation filtering}
\end{figure}

\paragraph {Innovation Filtering}: The innovation filtering discussed in Section \ref{IFISM} is ideal for noise spikes (abrupt  changes in sensor measurements) for short duration.  In Figure 11, a spike is injected into the EKF at $T = 125$ sec, with a duration of $0.3$ secs and the error is analyzed. The innovation filter NIS threshholding check (Eq. \ref{NIS_Threshold}) detects the spike and ignores the innovation in the Kalman Filter update step. Figures \ref{fig: EKF results without innovation filtering} and \ref{fig: EKF results with innovation filtering} show the ability of the Kalman Filter to ignore the spike and continue with accurate estimation.  


\vspace{1mm}

The experiment was repeated for the UKF and PF. Surprisingly, the UKF is unable to detect and filter small spikes but performs well with large ones (Figures \ref{fig: UKF results with innovation filtering, small spike} and \ref{fig: UKF results with innovation filtering, big spike}).


\begin{figure}
    \centering
    \includegraphics[width=0.70\linewidth]{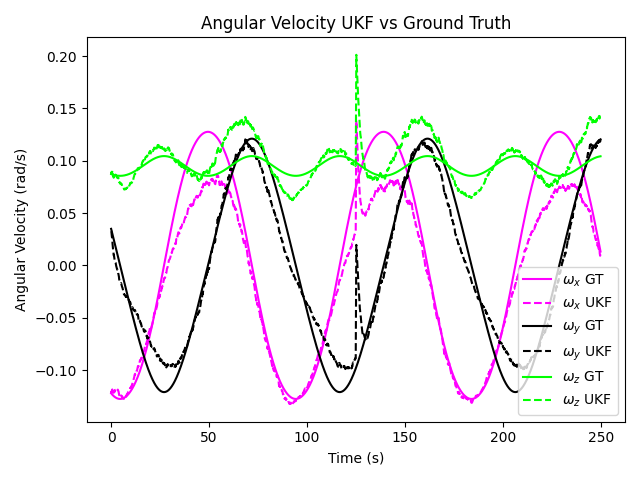}
    \caption{UKF Angular Velocity vs Time, with Innovation Filtering (Small Spike)}
    \label{fig: UKF results with innovation filtering, small spike}
\end{figure}

\begin{figure}
    \centering
    \includegraphics[width=0.70\linewidth]{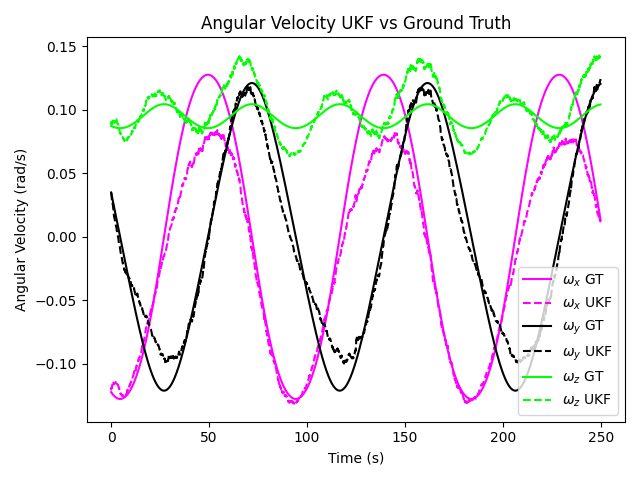}
    \caption{UKF Angular Velocity vs. Time, with Innovation Filtering (Large Spike)}
    \label{fig: UKF results with innovation filtering, big spike}
\end{figure}

This can be understood from Eqs. (\ref{eq:NIS_EKF}) and (\ref{sigma_yy}). As the non linearity causes the sigma points to spread out, and $\Sigma_{yy}$ becomes larger. This reduces NIS which reduces the sensitivy of the innovation filter. The unscented transform is designed to handle non-linearity but it also makes the filter more tolerant to smaller measurement spikes. Eq. (\ref{sigma_yy}) does provide some mitigations though. The measure noise covariance $R$ can be reduced to compensate when it's applicable.   



The PF with and without innovation-based fault rejection performs similarly (not shown), showing robustness to measurement spikes. This resilience stems from the PF’s sampling-based update: particles inconsistent with faulty measurements naturally receive low weights, reducing the impact of outliers without needing explicit rejection logic.

\begin{figure}
    \centering
    \includegraphics[width=0.70\linewidth]{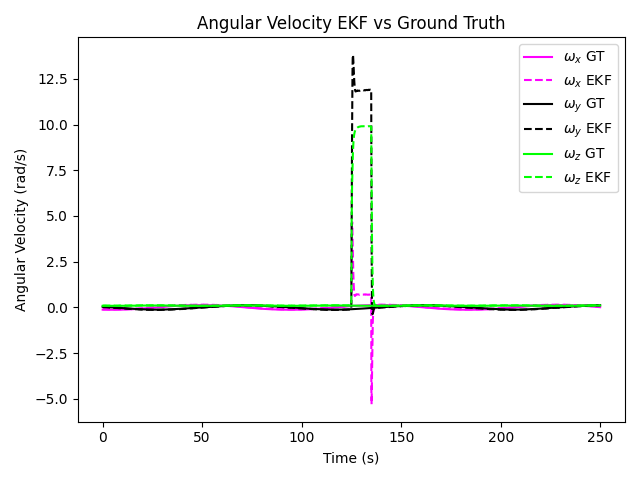}
    \caption{EKF Angular Velocity vs Time, without Innovation Sequence Monitoring}
    \label{fig:ekf_ism_no}
\end{figure}

\begin{figure}
    \centering
    \includegraphics[width=0.70\linewidth]{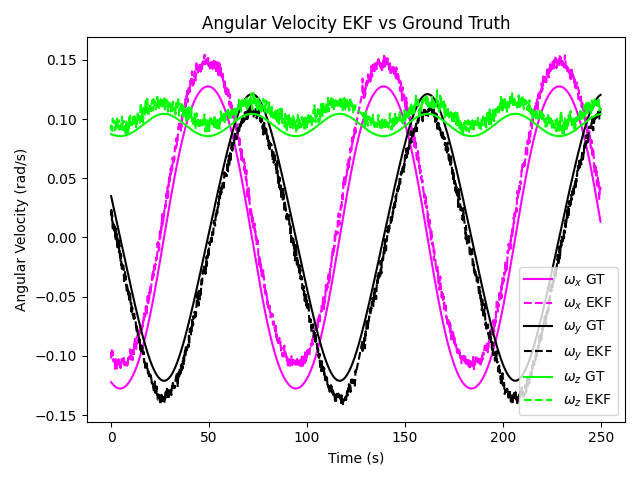}
    \caption{EKF Angular Velocity vs. Time, with Innovation Sequence Monitoring}
    \label{fig:ekf_ism_yes}
\end{figure}

\paragraph {Innovation Sequence Monitoring}: While the innovation filter detects short duration spikes, it performs poorly with intermittent dropouts. In Figures \ref{fig:ekf_ism_no} and \ref{fig:ekf_ism_yes}, a fault is injected into the EKF at $T = 125$ sec, but with a duration of $10$ sec. 
 Innovation sequence monitoring also discussed in Section \ref{IFISM} is used for these persistent anomalies. The innovation sequence filter NIS thresholding check (Eq. \ref{NIS_threshold_seq} detects the persistent anomaly and ignores the innovation in the Kalman Filter update step. Figure 16 shows the ability of the Kalman Filter to ignore the anomaly and continue with accurate estimation. This experiment is repeated for the UKF and the PF. It is observed that the detection performance of the PF has the highest performance followed by the UKF and EKF respectively.


\subsubsection{Sensor Fault Isolation}:  Fault isolation was introduced in Section \ref{FIIF} and summarized in Eq. \ref{IFISM}. It was found that the isolation mechanism provides much better sensitivity to short duration transient faults. Experiments were ran varying the magnitude of a noise spike for fault detection (overall NIS) and the fault isolation (individual sensor NIS). Through this, it was found that the isolator was able to detect and isolate smaller noise spikes. E.g., in one example a spike of $0.75$ $\frac{\text{rad}}{\sec}$ was added to the gyroscope. The detector failed to detect the spike. The detector required a minimum spike of $1$ $\frac{\text{rad}}{\sec}$ before it detected and filtered the spike. This corresponded to a threshold of $\chi^2_{11, \, 0.95}$ ($11$ is the dimensions of the measurement). The per sensor isolator was able to isolate the gyroscope as the source of error and detect a spike $0.75$ $\frac{\text{rad}}{\sec}$. This corresponded to a threshold of $\chi^2_{3, \, 0.95}$

Similar results for the innovation sequence monitoring are expected as it is based on a moving average of the same statistic.

\subsubsection{Sensor Fault Recovery}
\label{recovery}

\paragraph{Bias Estimation}

\begin{figure}
    \centering
    \includegraphics[width=0.70\linewidth]{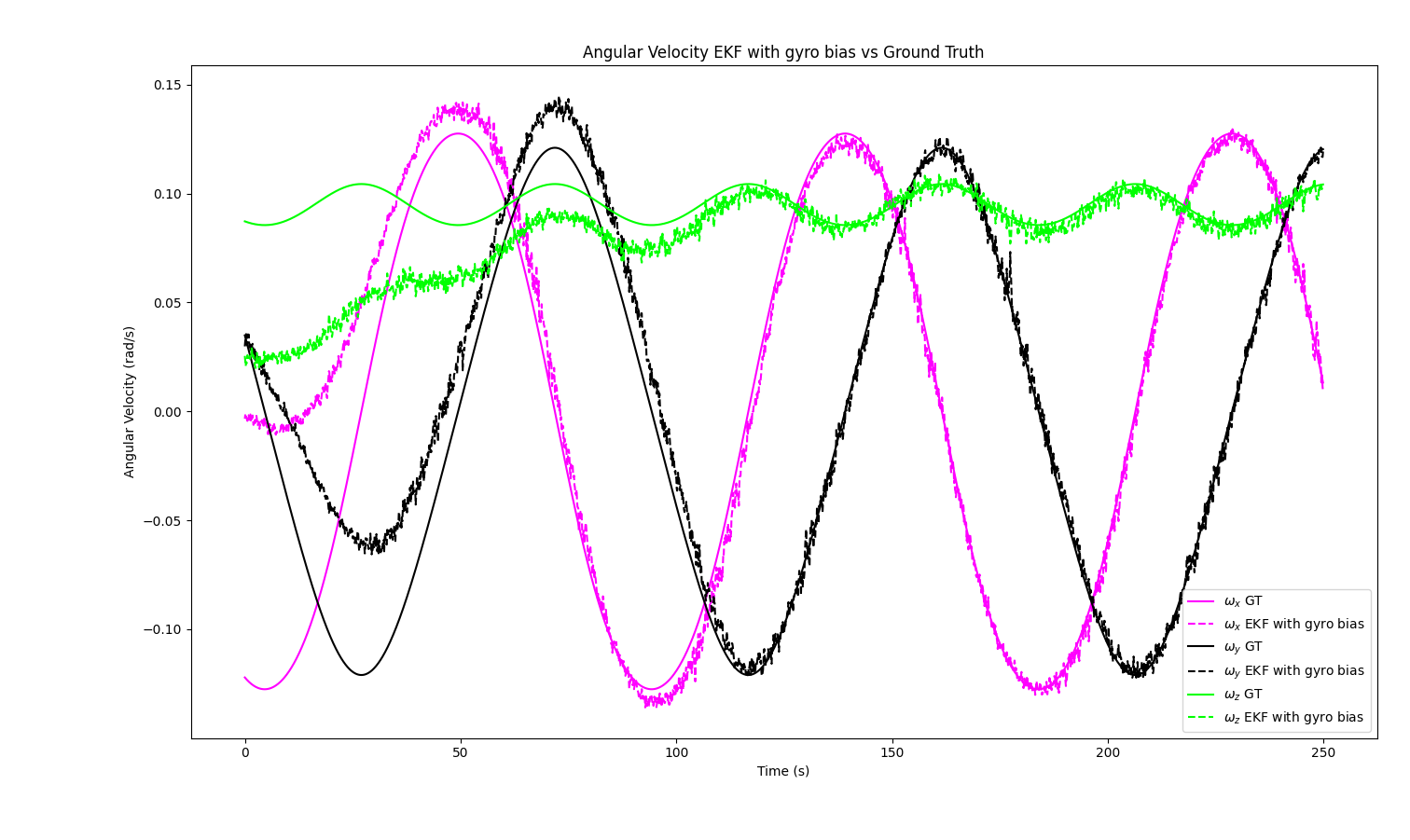}
    \caption{EKF Angular Velocity vs Time, Convergence of Gyro Bias Estimates}
    \label{fig:gyro_bias}
\end{figure}

Figure \ref{fig:gyro_bias} used the bias estimation model described in Section \ref{sec:recovery} to determine the bias. 

The estimate was able to correct the bias that was present throughout the angular velocity propagation in Figure \ref{fig:ekf_results_quats}. 



\begin{figure}
    \centering
    \includegraphics[width=0.70\linewidth]{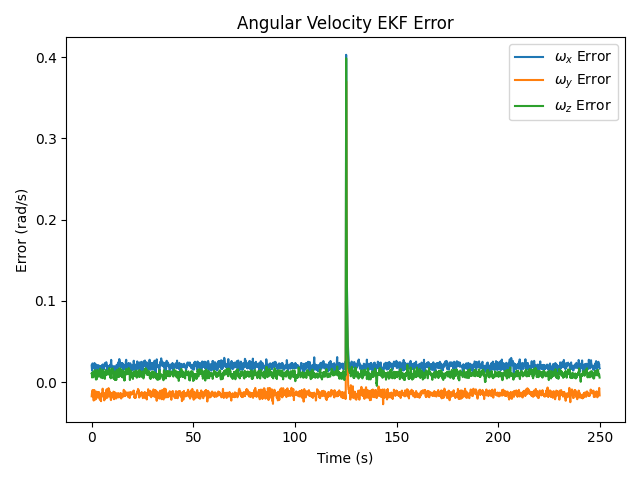}
    \caption{Angular Velocity Error vs. Time, without Sensor Fusion}
    \label{fig:sensor_fusion}
\end{figure}

\begin{figure}
    \centering
    \includegraphics[width=0.70\linewidth]{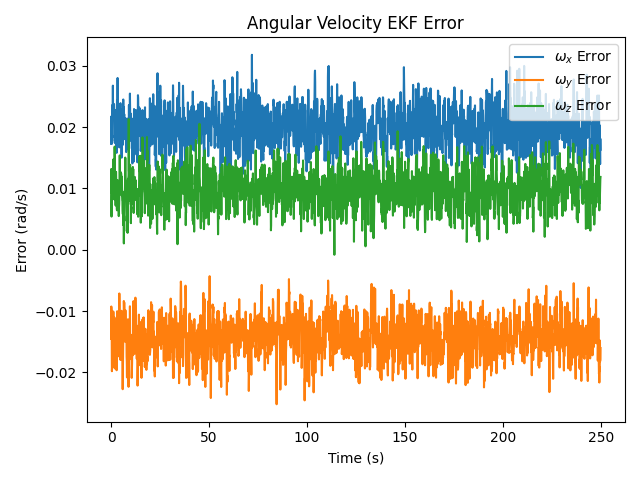}
    \caption{Angular Velocity Error vs. Time, with Redundant Sensor Fusion}
    \label{fig:redundant_fusion}
\end{figure}

\paragraph{Redundant Sensor Fusion}
Figures \ref{fig:sensor_fusion} and \ref{fig:redundant_fusion} used the redundant sensor fusion model  described in Section \ref{recovery} to disregard the faulty gyroscope and continue  the estimation using the star-tracker and the magnetometer. \uline{This is a remarkable result that shows the inherent fault tolerance of the Kalman Filter}

\section{Conclusions}
\label{conclusion}

For a large satellite in a low Earth orbit, the attitude and angular velocity were propagated by using Euler's equations for rigid body motion and computing the gravity gradient torque acting on the satellite. Applying the EKF, UKF, and PF to the model showed that all three filters were able to estimate the attitude of the spacecraft well with the EKF being the highest performing followed by the UKF and PF respectively. The EKF was unable to account for the bias in the gyroscope measurement, and the PF showed unstable results, making the UKF the highest performer.

The UKF, EKF, and PF provide a robust framework for state estimation in the presence of sensor faults. Faults including noise spikes, intermitent dropouts and biases were considered. The recursive form of the filters exhibits an inherent fault tolerance. This was used, in combination with techniques such as statistical analysis of  innovation residual, state vector augmentation, etc, to develop a systematic FDIR approach to sensor faults.

For fault detection, innovation filtering (statistical analysis of the innovation residual in update step) was applied to effectively detect and filter sharp transient faults like noise spikes. When comparing the performance of the filter in EKF, UKF, and PF implementations, surprisingly, the EKF performs better than the UKF and the PF. While UKFs and PF are more robust at handling non-linearities, modeling errors, and  non-Gaussian noise, EKFs are are better at filtering noise spikes.

Innovation filters perform poorly when dealing with intermittent dropout faults. Innovation sequence monitoring (statistical analysis of the moving average of the innovation residual) was applied to effectively filter these  faults. The innovation sequence monitor accumulates the effects of low amplitude sustained deviations - characteristic of intermittent step-type faults over time. They are better suited for persistent but subtle sensor degradation.

For fault isolation, innovation filtering was applied on a per-sensor basis. Using these statistics, faults can be attributed to specific sensors. This increases the sensitivity of the filter to lower magnitude noise spikes, which might otherwise be masked by aggregate statistics, calculated, across all sensors. Similar results are expected for innovation sequence monitoring, but the EKF was the focus in these experiments, given its superior performance. 

For fault recovery, the filters offers intrinsic support for redundant sensor fusion. Building on the fault isolation, the measurements models of the faulty sensor can be selectively excluded in the update step. The filter then seamlessly performs state estimation with the remaining sensors. Finally, by modeling sensor faults such as constant biases as part of the system state, the filters can estimate these values, and compensate for them.

\bibliographystyle{IEEEtran}
\bibliography{refs}

\appendix

All of the code for this project is on the Github repo: https://github.com/annahylbert05/aa273-final-project.

The main branch includes all of the sim setup and standard filter implementation. The branch innovation\_filter included all of the FDIR work that was done.

\end{document}